# Aerial Rock Fragmentation Analysis in Low-Light Condition Using UAV Technology

Thomas Bamford[1]*, Kamran Esmaeili[1], Angela P. Schoellig[2]
[1] *Lassonde Institute of Mining, University of Toronto, Canada;* [2] *University of Toronto Institute for Aerospace Studies, Canada*
*Corresponding author: thomas.bamford@mail.utoronto.ca

In recent years, Unmanned Aerial Vehicle (UAV) technology has been introduced into the mining industry to conduct terrain surveying. This work investigates the application of UAVs with artificial lighting for measurement of rock fragmentation under poor lighting conditions, representing night shifts in surface mines or working conditions in underground mines. The study relies on indoor and outdoor experiments for rock fragmentation analysis using a quadrotor UAV. Comparison of the rock size distributions in both cases show that adequate artificial lighting enables similar accuracy to ideal lighting conditions.

**Introduction**

In recent years, Unmanned Aerial Vehicle (UAV) technology has been introduced to the mining industry to conduct terrain surveying, monitoring and volume calculation tasks. These tasks are essential for mining operations, but they do not leverage all the benefits that UAVs can offer to the industry. In general, UAVs can be used for continuous acquisition of high-resolution data, which can be beneficial in blast design, monitoring mine operations and other mine-to-mill optimization campaigns, enabling more efficient and faster response to changes in the mining process conditions. Moreover, compared to manual techniques, data acquisition with UAVs can be automated to provide higher spatial- and temporal-resolution data which, in turn, improves the statistical reliability of measurements. Other benefits of UAV-based data collection in the minerals industry include: no disruption of production, safety of technicians, and being able to collect data from typically inaccessible and hazardous areas.

Rock breakage by drilling and blasting is the first phase of the production cycle in most mining operations. Measurement of post-blast rock fragmentation is important because the rock size distribution can greatly influence the efficiency of all downstream mining and comminution processes [1]. Studies have shown that rock fragmentation can influence the volumetric and packing properties of rock (e.g. the fill factor and bulk volume) and, consequently, the efficiency of digging and hauling equipment [1]. Similarly, other studies have demonstrated the direct influence of the rock size distribution on comminution energy consumption, mill throughput rates and the productivity of these processes [2]. Blasting engineers also use the rock size distribution as a means of quality control of blast design and operation. Due to these impacts, the continuous measurement of post-blast rock fragmentation is important to the optimization of a mining operation.

Several methods have been developed for estimating rock size distribution. These include visual observation by an expert, sieve analysis, and 2D and 3D image analysis [3]. Visual observation involves inspecting the rock pile and subjectively judging the quality of the blasting material. This subjective method often produces inaccurate and imprecise results. Sieve analysis, or screening, involves passing a sample of the rock pile through a series of different sieve size trays. This method generates more consistent and accurate results; however, it is more expensive, time consuming, and in certain cases, impractical to perform.

Image analysis techniques for measuring rock fragmentation are commonly used in modern mining operations because they enable practical, fast, and relatively accurate measurements [4]. Different sensors and approaches can be used for image collection and processing [5]. Recent techniques include using Light Detection and Ranging (LiDAR) sensors [6] and Deep Neural Network image segmentation [7]. However, the most common technique is to capture images of a post-blast muck pile from fixed ground locations, using a monocular camera and scale objects. This method, referred to as the conventional method, involves a technician walking to a rock pile, placing scale objects of known size in the region of interest, and capturing individual 2D images. Sanchidrián et al. [4] discussed several limitations of the conventional method for rock fragmentation analysis. Among these limitations, image system resolution is of high importance, as it can lead to inaccurate rock size measurement.

Using 3D measurement techniques to capture images of rock piles has overcome some of these limitations, [6, 8]. Application of 3D imaging methods eliminates the need for placement of scale objects and reduces the error that is created by the uneven shape of the rock pile. Using LiDAR sensors for rock fragmentation analysis can reduce the delineation error produced by uneven shape of the muck pile and the resolutions error associated with poor lighting condition [8]. While the 3D techniques improve image analysis methods, there are still limiting aspects. One example of this is the significant amount of time required to capture detailed scans with LiDAR technology [6]. Another drawback of the 3D imaging techniques is that they are limited to capturing images from fixed locations because motion blur can significantly smooth out the 3D data, making particle delineation difficult [8].

To improve the spatial and temporal resolution of image analysis for measurement of rock size distribution and to automate the data collection process, our work has focused on using UAVs to conduct aerial rock fragmentation analysis [9, 10, 11]. While UAVs can be configured to carry LiDAR systems, which may reduce delineation error in poor lighting conditions, current commercially available UAV systems are configured with high-resolution monocular cameras. As a result, if the UAV system captures images in poor lighting conditions, boundaries of rock fragments are difficult to delineate and/or particles become obstructed by shadow and darkness. These lighting conditions represent night shifts in surface mines and working conditions in underground mines. Enabling UAV measurement in these conditions is essential for automated and continuous post-blast rock fragmentation analysis, since mining environments rarely provide ideal conditions for image-based methods.

This work investigates the application of UAVs for measurement of rock fragmentation under poor lighting conditions. The study relies on indoor and outdoor experiments using a quadrotor UAV. First, a rock pile with known size distribution was photographed by the UAV in a lab. The experiments were carried out in both ideal lighting and dark conditions. For the dark case, different artificial lighting, separate from the UAV, were used to illuminate the rock pile. The same experiments were conducted outdoors. Comparison of the rock size distributions from both experiments show that adequate and evenly distributed artificial lighting allows producing similar accuracy to the ideal lighting condition.

## Experiment Setup
*Aerial Vehicle System*

The components and overall configuration of the aerial vehicle system used in this study for rock fragmentation measurement is illustrated in *Figure 1*. The system is similar to what was employed in [9, 11], but differs as a component for artificial lighting was added.

**Unmanned Aerial Vehicle.** A commercially available UAV with integrated camera, the Parrot Bebop 2, was used in our experiments because it has onboard image stabilization and a Global Positioning System (GPS) receiver. A list of the main specifications and performance of the vehicle can be found in [12]. This UAV can capture stabilized high-resolution photos and videos, which is essential for accurate image analysis. The GPS receiver provides position measurements for vehicle control. In these tests, the UAV broadcasted video streams with a resolution of $856 \times 480$ pixels stabilized onboard during flight. The camera orientation was stabilized onboard by moving a virtual window through the field of view of the integrated fisheye lens. The UAV receives camera commands and transmits the camera orientation in tilt and pan.

**Software Framework.** The open-source Robot Operating System (ROS) [13] was used to act as the central software framework of the aerial vehicle system (see *Figure 1*). For the proposed system, ROS wirelessly sent high-level flight commands to the UAV to follow a user-defined flight plan, which required the use of onboard position and orientation measurements for feedback. Images captured from the UAV video stream were either stored onboard the UAV for offline analysis or off-board on the ground control station for real-time analysis. Images stored for real-time processing were analyzed in a standard fragmentation analysis software, Split-Desktop [14]. We used a keyboard and mouse macro to run the image analysis automatically.

Since flight safety is of paramount importance, three safety measures were implemented into the system. By default, a safety pilot could take control over the UAV using a joystick, the Xbox Wireless Controller, that was interfaced with ROS. In case of joystick malfunction, a ROS graphical user interface could have been used to initiate emergency procedures on the ground control station. This user interface was also used to display flight information on the ground control station. Finally, if connection with the UAV was lost, the drone would automatically fly to its predefined home position.

**Rock Fragmentation Analysis.** For image analysis, Split-Desktop [14], an industry standard software for fragmentation analysis in mining, was used. The software receives an image and delineates particles using image segmentation. Scale objects are traced graphically to set the image scale. This assumes that (1) the scale object lies on the rock pile surface, and (2) the surface is planar. To avoid using physical scaling objects, as described in [10], a point cloud created by the aerial vehicle system could be used to compute image scale. In this work, physical scale objects were used to determine image scale.

**Flight Plan.** For aerial rock fragmentation analysis, flight plans were created to capture nine images at a camera tilt angle of 83 degrees, for a fixed altitude of 0.5 m above the rock pile while ensuring no or little image overlap. The tilt angle chosen is the maximum that the UAV specification allowed. The images were taken approximately perpendicular to the rock pile surface, as suggested by the fragmentation analysis software [14].

**Light Source.** To illuminate the rock pile during fragmentation measurement, natural and artificial lighting were used. In these experiments, we used a variety of lighting conditions with different light sources. Fixed artificial lighting sources (not attached to the UAV) were setup for dark conditions. The amount of luminous emittance from the light sources was measured in luminous flux per unit area (with the SI unit lux). We also measured the illuminance (in lux) of the subject rock pile.

*Indoor Configuration*

The aerial vehicle system used in the indoor experiments differs from the outdoor experiments in the global positioning system used. The lab is equipped with a motion capture camera system for precise UAV localization and control. This commercially available system uses ten 4-megapixel Vicon MX-F40 cameras and reflective markers attached to each subject to measure position and orientation at a rate of 200 Hz. For the lab experiments, the position and orientation of the rock pile and the UAV are collected and sent to ROS to control the motion of the UAV relative to the pile. The indoor lab has fluorescent lighting and is free of wind.

A pile of rock fragments with different sizes, ranging from coarse gravel (19 mm) to fine sand (<4 mm), was built in the lab. Prior to forming the pile, the rock fragments were put through sieve analysis to determine the 'true' rock size distribution as a reference for computing measurement accuracy. The results of the sieve analysis are presented for four discrete screen sizes, which is referred as the reference curve in the "Results" section. To use the sieve analysis as a reference for sizes greater than 19 mm, the Swebrec rock size distribution [4] curve was fit to the collected data. The parameters of this distribution are presented in [9]. Spherical scale objects, with a diameter of 60 mm, were used to provide image scale for fragmentation analysis in the indoor experiments.

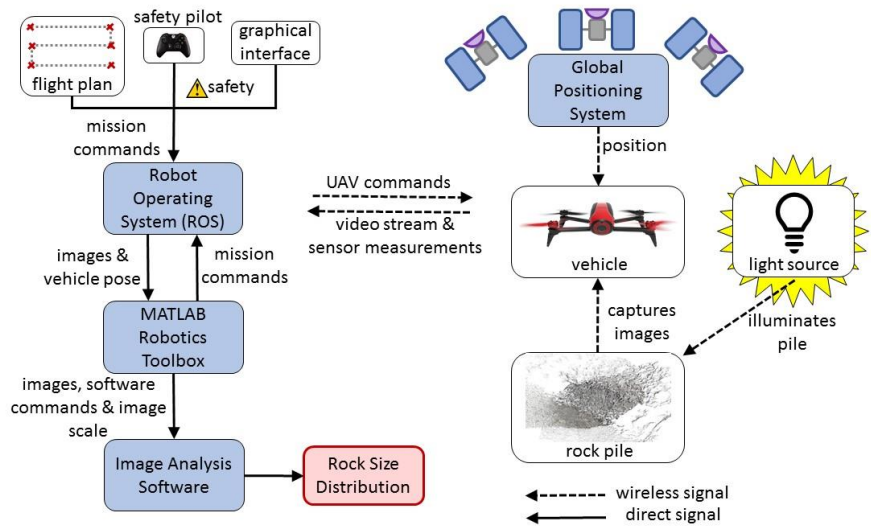

Figure 1: Block diagram of the aerial vehicle system used for aerial fragmentation analysis. Arrows show the typical information flow while conducting a flight plan.

*Outdoor Configuration*

Outdoor experiments were conducted in the outdoor robotics experiment area at the University of Toronto. The aerial vehicle system used in this configuration employs a GPS receiver for UAV localization and control. The pile of rock fragments, used in the indoor configuration, was measured in variable lighting conditions outdoor. The 'true' rock size distribution was used as a reference for computing measurement accuracy. The same spherical scale objects, used in the indoor configuration, were used to provide image scale for fragmentation analysis in the outdoor configuration.

**Methodology**

To investigate the application of artificial lighting for measurement of rock fragmentation under poor lighting conditions, indoor and outdoor experiments were carried out using the aerial vehicle system described above. Steps were taken in the indoor and outdoor experiments to ensure a fair comparison of results under different lighting conditions. In both experiments, the same flight plans were flown by the aerial vehicle system while ensuring that the rock pile configuration was fixed.

The procedure used to conduct indoor experiments was:

1. Measure points around rock pile base using motion capture camera system;
2. Place scale objects on the rock pile;
3. Configure artificial lighting system, if required;
4. Record the luminous emittance of the light source;
5. Record the illuminance of the rock pile;
6. Capture images following the rock fragmentation analysis flight plan;
7. Transfer images to workstation; and
8. Conduct rock fragmentation measurement with image analysis software.

The procedure used to conduct outdoor experiments differs from the indoor procedure in step 1, where instead of using the motion capture camera system to measure points around the rock pile base a GPS receiver was used.

**Results**

The following subsections present results for the indoor and outdoor experiments, respectively. These results compare the rock size distributions measured by the aerial vehicle system in a variety of lighting conditions.

*Lab Experiments*

A variety of lighting conditions were created in the lab environment, as described in *Table 1*. Poor lighting conditions (120 lx, 40 lx, and 11 lx) were created to represent night shifts in surface mines or working conditions in underground mines (3 lx in stopes and 7.5 lx in haulages [15]). Then, using the procedures described above, aerial rock fragmentation analysis was conducted for each condition. *Figure 2* provides examples of the images collected in ideal (Experiment #1) and dark (Experiment #4) lighting conditions alongside the delineation net produced. As can be seen, the delineation created in the ideal case is more accurate than those created in the dark case because particle boundaries are difficult to identify.

To determine prediction accuracy for each lighting condition, the percent error residuals for percent passing with respect to the reference sieve analysis curve were computed for the discrete sieve series. Percent error residuals are calculated for percent passing ($P(< x)$) according to:

$$\text{Percent error residuals} = \frac{P(< x)_{\text{Image Analysis}} - P(< x)_{\text{Sieve Analysis}}}{P(< x)_{\text{Sieve Analysis}}} \times 100\%$$

where $P(< x)_{\text{Image Analysis}}$ is the percent passing a size of $x$ measured by the image analysis software and $P(< x)_{\text{Sieve Analysis}}$ is the percent passing a size of $x$ measured by the sieve analysis. The rock fragmentation analysis for each experiment, with residuals, is plotted in *Figure 3*. The Swebrec rock size distribution function [4] was fit to the discrete points from sieve analysis and image analysis methods so that a 2-norm error between the image analysis and the reference curve could be calculated. *Figure 4* presents the illuminance measured for each experiment against the 2-norm error.

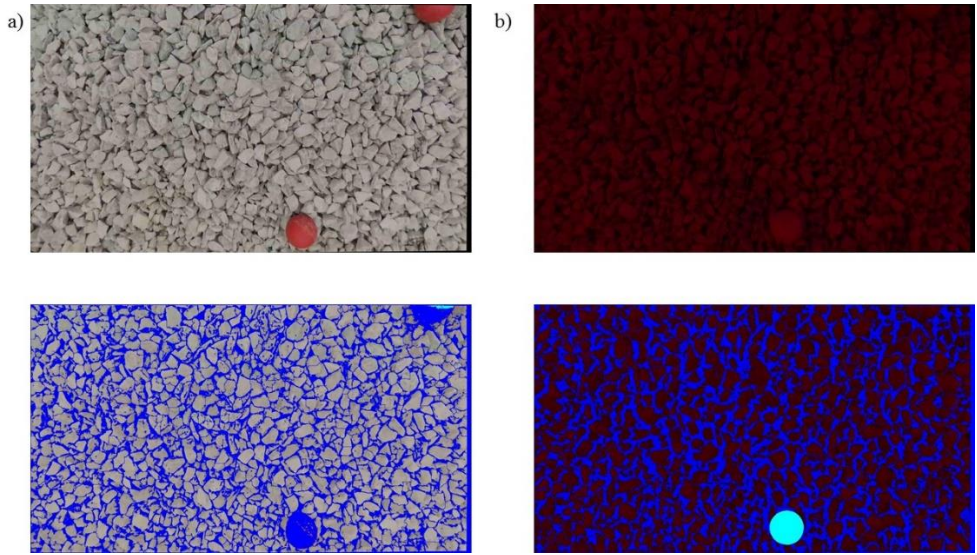

Figure 2: a) Raw and delineated photo captured in ideal lighting (450 lx). b) Raw and delineated photo captured in dark conditions (11 lx).

**TABLE 1 Indoor experiment lighting conditions**

| Experiment Number | Experiment Description | Illuminance of the Rock Pile (lx) | Luminous Emittance (lx) | Position of Light Source |
|---|---|---|---|---|
| 1 | Normal lighting | 450 | 1815 per light | ceiling, 3 m above pile |
| 2 | Dim lighting | 120 | 1815 per light | ceiling, 3 m above pile |
| 3 | Uneven lighting | 40 | 1815 per light | ceiling, 3 m above pile |
| 4 | Dark | 11 | NA* | NA |
| 5 | Artificial lighting 1 | 14 | 1000 | 20° tilt, 3 m from pile center |
| 6 | Artificial lighting 2 | 18 | 1000 | 30° tilt, 2 m from pile center |

*NA = not applicable.

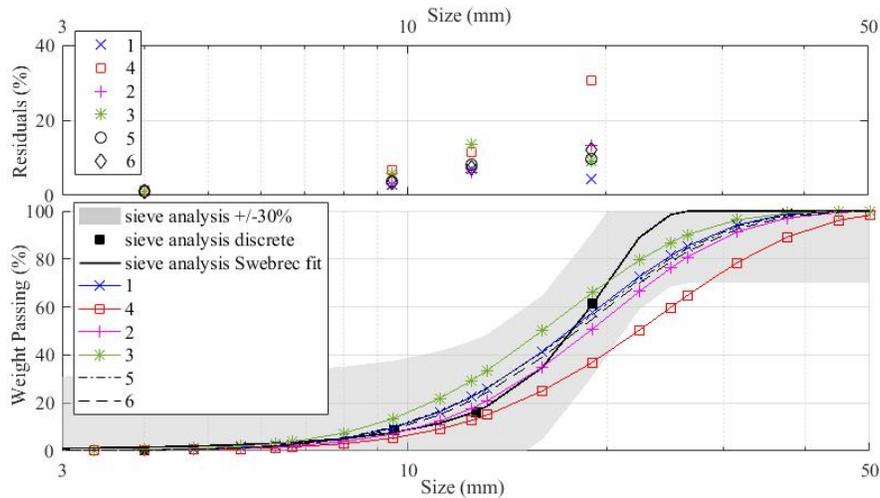

Figure 3: Rock fragmentation analysis results for indoor experiments with respect to the sieve analysis reference curve. Discrete points represent the combined results for each set of images collected during each experiment. The gray envelope represents the accepted maximum error envelope of 30% recommended by [4] for industry standard 2D image analysis.

The rock size distribution estimated in the normal lighting case (450 lx) is shown to remain within the 30% maximum error envelope suggested by [4]. In the dark case (11 lx), the estimated distribution is far from the true distribution and falls out of the maximum error envelope, as was expected. In the uneven lighting case (40 lx), the error increased because a combination of shadows and bright particle faces created particle disintegration and fusion. To ensure that this was avoided in artificial lighting conditions, light was applied evenly. In the artificial lighting cases (14 lx and 18 lx) and dim lighting (120 lx), the predicted distributions remained in the acceptance envelope. The artificial lighting conditions produced distributions that were very close to the distribution predicted in the ideal conditions. The results indicate that artificial lighting, even if it marginally increases the illuminance of the rock pile, enables accurate prediction of rock fragmentation.

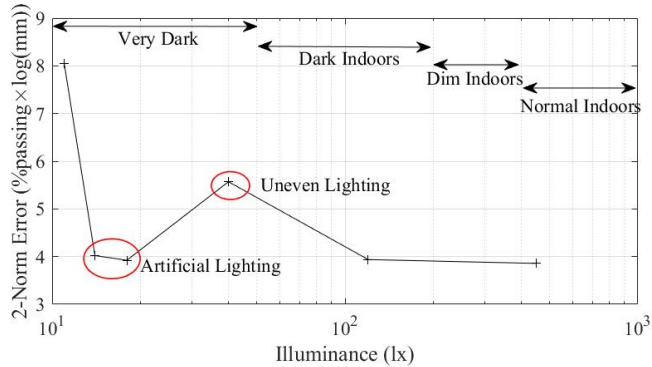

Figure 4: 2-norm error between the estimated rock size distributions and the sieve analysis reference curve plotted against illuminance measurements for indoor experiments. Qualitative measures of common lighting conditions are provided as ranges of illuminance measurements.

*Outdoor Experiments*
Having shown that applying artificial lighting in poor lighting conditions improves the accuracy of results indoors, we applied this method in outdoor experiments at the site specified above. We also conducted aerial rock fragmentation analysis in a variety of lighting conditions to investigate the effect lighting has on prediction accuracy, as described in *Table 2*. Using the procedures described above, aerial rock fragmentation analysis was conducted for cloudy, dusk, and dark conditions as well as using artificial lighting to illuminate the pile in dark conditions. The rock fragmentation analysis for each experiment, with residuals, is plotted in *Figure 5*. *Figure 6* presents the illuminance measured for each experiment against the 2-norm error.

The rock size distributions estimated in the cloudy (9500 lx), and dusk (363 lx) conditions are shown to remain within the 30% maximum error envelope suggested by [4]. In the dark case (3 lx), the UAV system was not able to measure fragmentation of the rock pile because the camera did not measure any light. In the artificial lighting cases (14 lx), the predicted distributions remained in the acceptance envelope. The artificial lighting condition produced a distribution that was close to the distribution predicted in the cloudy conditions. These results are very promising because an even artificial lighting has enabled accurate prediction of rock fragmentation in outdoor settings.

**TABLE 2 Outdoor experiment lighting conditions**

| Experiment Number | Experiment Description | Illuminance of the Rock Pile (lx) | Luminous Emittance (lx) | Position of Light Source |
| --- | --- | --- | --- | --- |
| 1 | Cloudy | 9500 | 54000 | NA[*] |
| 2 | Dusk | 363 | 1966 | NA |
| 3 | Artificial lighting | 14 | 18 | 45° tilt, 30 m from pile |
| 4 | Dark | 3 | 3 | NA |

[*]NA = not applicable.

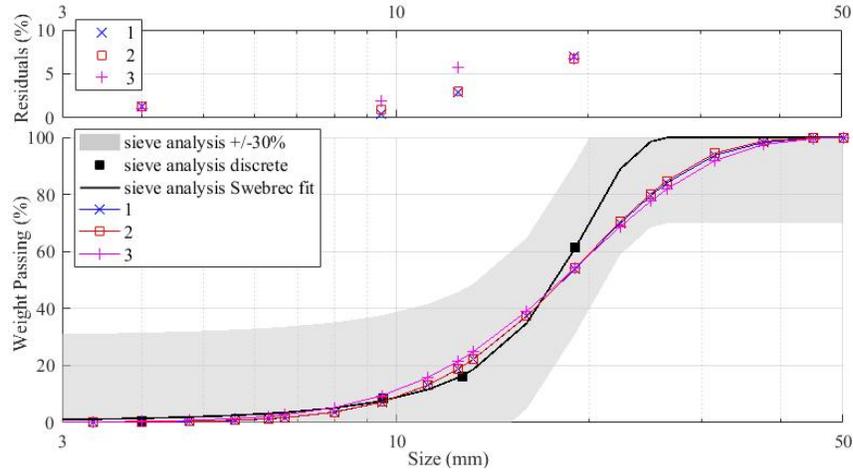

Figure 5: Rock fragmentation analysis results for lab experiments with respect to the sieve analysis reference curve. Discrete points represent the combined results for each set of images collected during each experiment. The gray envelope represents the accepted maximum error envelope of 30% recommended by [4] for industry standard 2D image analysis.

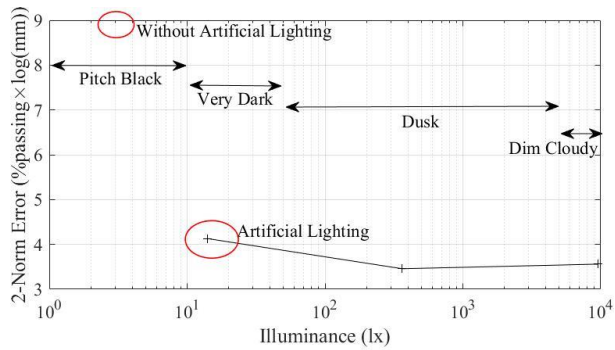

Figure 6: 2-norm error between the estimated rock size distributions and the sieve analysis reference curve plotted against illuminance measurements for outdoor experiments. Qualitative measures of common lighting conditions are provided as ranges of illuminance measurements.

**Conclusion**

This work presented the results of indoor and outdoor experiments to investigate the effect of lighting conditions on aerial rock fragmentation analysis. The use of fixed artificial lighting to collect images in poor lighting conditions was proposed to improve accuracy of rock fragmentation analysis using UAV. Results, from both the indoor and outdoor experiments, show that lighting conditions greatly impact the accuracy of image analysis techniques for rock fragmentation measurement. Applying the artificial lighting evenly to the rock pile, can improve prediction accuracy. In this study, fixed artificial lighting was used for the experiments, however, in hazardous or inaccessible areas like underground stopes, the artificial lighting can be attached to the UAV system or another vehicle can be used for the artificial lighting. Continuous measurement of the rock size distribution is important at different lighting conditions because the rock size distribution can change while the muck pile is excavated, which impacts the potential efficiency of downstream processes. Aerial fragmentation analysis shows promise to be a faster, more accurate, and high resolution measurement technique for the mining industry, and with the addition of artificial lighting systems poor lighting conditions can be mitigated.

**Acknowledgement**

This work was supported by Split-Engineering, the University of Toronto's Dean's Strategic Fund, the Canada Foundation for Innovation John R. Evans Leaders Fund, and the Natural Sciences and Engineering Research Council of Canada.